\documentclass[letterpaper, 10 pt, conference]{ieeeconf}  

\IEEEoverridecommandlockouts                              

\pdfoutput=1

\usepackage{graphicx} 
\usepackage{color}
\usepackage[ruled,linesnumbered]{algorithm2e}
\graphicspath{{figures/}}
\usepackage{subcaption}
\usepackage{hyperref}
\usepackage{amsmath,amsfonts}

\title{\LARGE \bf
Learning Robotic Assembly from CAD
}

\author{Garrett Thomas*$^{1}$, Melissa Chien*$^{1}$, Aviv Tamar$^1$, Juan Aparicio Ojea$^2$, Pieter Abbeel$^1$
\thanks{
*Equal Contribution
$^1$EECS Department, UC Berkeley
$^2$Siemens Corporation}%
}

\newcommand{\loss}{\ell}
\newcommand{\FK}{f_\textrm{FK}}
\newcommand{\cs}{\phi}
\newcommand{\ts}{x}
\newcommand{\robot}{\mathcal{A}}
\newcommand{\obstacles}{\mathcal{O}}
\newcommand{\cspace}{\mathcal{C}}
\newcommand{\cfree}{\cspace_{free}}
\newcommand{\ttraj}{T_{\textrm{traj}}}
\newcommand{\tfinal}{T_{\textrm{final}}}
\newcommand{\rtraj}{\ts^{\textrm{ref}}}
\newcommand{\lossmp}{\ell^{\textrm{ref}}}

\begin{document}

\maketitle
\thispagestyle{empty}
\pagestyle{empty}

\begin{abstract}
In this work, motivated by recent manufacturing trends, 
we investigate autonomous robotic assembly.
Industrial assembly tasks require contact-rich manipulation skills, which are challenging to acquire using classical control and motion planning approaches. Consequently, robot controllers for assembly domains are presently engineered to solve a particular task, and cannot easily handle variations in the product or environment. Reinforcement learning (RL) is a promising approach for autonomously acquiring robot skills that involve contact-rich dynamics. However, RL relies on random exploration for learning a control policy, which requires many robot executions, and often gets trapped in locally suboptimal solutions. Instead, we posit that prior knowledge, when available, can improve RL performance. We exploit the fact that in modern assembly domains, geometric information about the task is readily available via the CAD design files. We propose to leverage this prior knowledge by guiding RL along a geometric motion plan, calculated using the CAD data. We show that our approach effectively improves over traditional control approaches for tracking the motion plan, and can solve assembly tasks that require high precision, even without accurate state estimation. In addition, we propose a neural network architecture that can learn to track the motion plan, thereby generalizing the assembly controller to changes in the object positions.
\end{abstract}

\section{INTRODUCTION}
\label{sec:intro}


A dominant trend in manufacturing is the move toward small production volumes and high product variability \cite{lasi2014industry}. 
It is thus anticipated that future manufacturing automation systems will be characterized by a high degree of autonomy, and will be able to
learn new behaviors without explicit programming. In this work, we look at an important problem in manufacturing: the assembly of mechanical objects.

Robotic assembly typically involves object manipulation tasks with substantial contacts and friction, such as inserting or removing tight-fitting objects, or twisting a bolt into place. Designing robot controllers for such tasks is difficult because of the complexity of modelling and estimating contact dynamics accurately. Consequently, nearly all real-world robotic assembly applications are implemented in repetitive scenarios, which can pay off the substantial engineering efforts required~\cite{durkop2015analyzing}. In addition, the implementations often rely on clever (special-purpose) fixtures to guide the assembly and part feeders for assuring repetitive initial conditions. 

Prominent approaches for autonomous manipulation are based on either motion planning \cite{lavalle2006planning,latombe2012robot,donald1993kinodynamic}, or reinforcement learning (RL) \cite{deisenroth2013survey,kober2013reinforcement,levine2015learning}. In motion planning, geometric and dynamic knowledge about the task is used to plan a trajectory for the robot, which is then executed using a tracking controller.

\begin{figure}
    \centering
    \includegraphics[width=\linewidth]{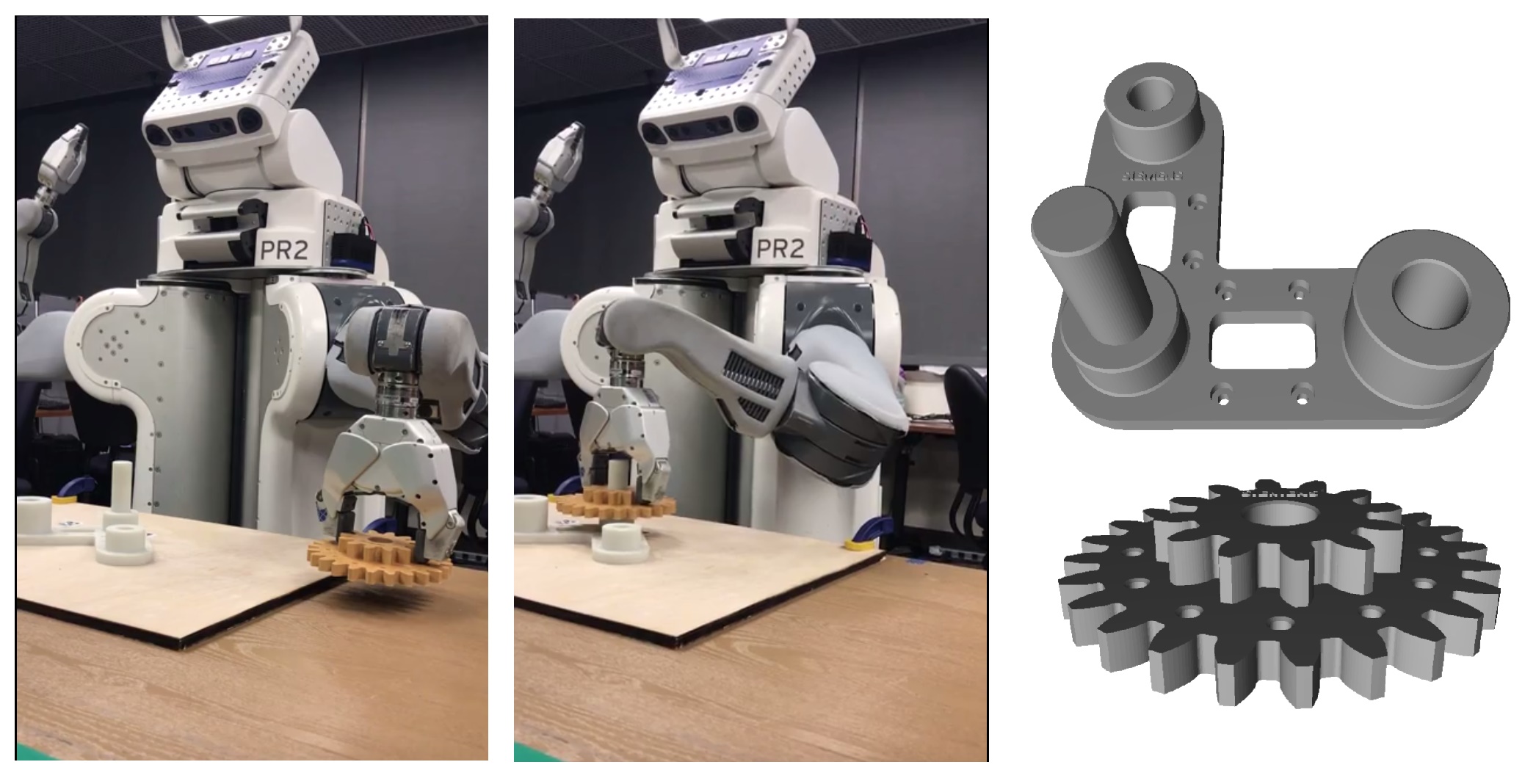}
    \caption{Our method learns to assemble objects specified in CAD files. The CAD data is used to compute a geometric motion plan, which is then used as a reference trajectory for a reinforcement learning algorithm on the real robot.}
    \label{fig:example}
\end{figure}
In contact rich environments such as robotic assembly, it is required that either the controllers or the modeling of contact and friction dynamics are highly accurate, both of which can be difficult to obtain in practice. In RL, on the other hand, no prior knowledge about the dynamics is required\footnote{This distinguishes RL from the traditional optimal control setting.}, and the task is specified only by means of a scalar cost function, such as a distance to a goal. A controller for the task is learned through interaction with the system by searching for a reactive policy that minimizes the cost.

While RL has shown promise in learning policies for contact-rich manipulation \cite{levine2015learning,fu2016one,gu2017deep}, the policy search can converge to undesirable local minima in the cost, and a reactive policy can fail to generalize to simple modifications of the task such as moving the object positions \cite{Tamar16}. Furthermore, without prior knowledge to guide the search, RL methods tend to suffer from high sample complexity, which further challenges their use in real-world robotic applications.

In this work we propose a method that combines motion planning with RL policy search for efficient learning of assembly tasks. We exploit the fact that in most real-world assembly domains, high quality geometric information about the objects is readily available in the form of CAD models used to manufacture the assembly parts. Thus we can obtain a geometric motion plan for the task and use it as guidance for the RL algorithm. 
Our approach has two main ideas: 
\begin{enumerate}
    \item A cost function for RL that `tracks' the motion plan, to effectively avoid locally optimal solutions.
    \item A neural-network policy representation that takes as input both an observation and a motion plan, for generalizing to task modifications.
\end{enumerate}
We show that our approach can efficiently learn to assemble mechanical objects using the PR2 robot by exploiting a CAD specification of the task. Our resulting policy generalizes to modifications in the task, such as the object placements. In addition, our approach succeeds in tasks where noisy state estimation and local minima cause standard motion planning and RL approaches to fail. Finally, the prior knowledge we exploit leads to a fast convergence of the RL algorithm, further motivating our approach for real-world domains.


\section{RELATED WORK}
\label{sec:related}
Various aspects of autonomous assembly have been explored in the robotics community. Early works such as \cite{de1991correct} and \cite{halperin2000general} explored how to transpose a CAD file to a sequence of assembly steps. Recently, \cite{knepper2013ikeabot} extended this approach to autonomous furniture assembly, and~\cite{heger2010assembly} explored extensions for error detection and recovery. In these works, the focus is on \emph{planning} -- inferring the correct sequence of assembly steps, while the execution is performed using special purpose hardware for each task (e.g., the special purpose grippers in~\cite{knepper2013ikeabot}). In this work we dismiss the planning component in assembly, and consider simple tasks that involve only two parts. Our focus, however, is on the \emph{control} part, and we investigate learning controllers on a general purpose platform such as the PR2 robot.

Combining RL with a reference trajectory has been explored in learning with dynamic movement primitives \cite{peters2008reinforcement,pastor2009learning,mulling2013learning}. In these works, the reference trajectory is obtained from a human demonstration, and represented parametrically, using supervised learning. RL is then used to modify the trajectory parameters, and the trajectory is tracked using a fixed controller. Our work acquires the trajectory from motion planning and does not require any human demonstration. Instead of modifying the reference trajectory, we learn to modify the controller that tracks it.

Given a fixed motion plan, iterative learning control (ILC)~\cite{moore2000iterative,longman2000iterative,bristow2006survey,wang2009survey} sequentially adapts a controller to track it. The iLQR trajectory optimization stage in our algorithm is essentially a form of ILC; however, we also learn a neural network policy that \emph{generalizes} the learned controller to different motion plans.

The recent work of Duan et al.~\cite{duan2017one} proposed an imitation learning method that trains a NN policy to take as input a reference demonstration, using soft attention. Our NN representation is similar in spirit and applies a similar soft attention mechanism to the motion plan. 

Our cost function for tracking the motion plan can be seen as a form of reward shaping~\cite{ng1999policy}. Designing reward functions for RL is a difficult task and often approached using additional human feedback, e.g. inverse RL and preference based RL~\cite{ng2000algorithms,abbeel2004apprenticeship,ziebart2008maximum,wirth2016model,finn2016guided}. In contrast, we exploit prior geometric information about the task for reward design.

Kinodynamic motion planning \cite{donald1993kinodynamic,hsu2002randomized,karaman2010optimal,xie2015toward} can plan motions that account for the dynamics in the task. However, it requires knowing the dynamics model in advance, which is difficult to obtain in the contact-rich manipulation scenarios we consider. Instead, we focus on \emph{geometric} motion planning, and handle the dynamics using RL.

Finally, control using force/torque sensing is an effective method for assembly tasks~\cite{lange2012control,lange2013force}. In this work our controller is based on position and velocity input, though our method can be directly extended to use additional sensory modalities such as force/torque.


\section{PRELIMINARIES}
\label{sec:background}

\subsection{Problem Formulation}\label{ssec:problem_formulation}

We consider robot manipulation tasks that can be described as moving a grasped object to a goal position. This family of problems includes common household and industrial manipulation tasks, such as assembling and disassembling parts together, inserting and ejecting objects, and movement in high-friction domains. 
We assume that a collision-free trajectory to the goal exists, in the sense that it can be computed using motion planning. However, we make no particular assumptions on the \emph{dynamics} experienced throughout the task execution, and in particular, do not restrict contacts and friction. 

Let $\cs_t$ and $\ts_t$ denote the state of the robot in configuration space (joint space) and task space (end-effector pose), respectively, at time $t$, and let $u_t$ denote the control command (torque) applied at that time. Given a goal state in task space $\ts_g$, an initial robot state $\cs_0$, and a time horizon $T$, our manipulation problem is formulated as\footnote{It is straightforward to extend this formulation to stochastic dynamics, by considering the expected loss, e.g., as in \cite{levine2015learning}. It is also straightforward to add to the cost constraints on velocity or time dependence. For clarity, we restrict our presentation to this simple and deterministic setting.}
\begin{align}\label{eq:prob_def}
    \min_{u_0,\dots,u_{T-1}} & \loss(\ts_T, \ts_g), \\ \nonumber
    \textrm{s.t. } & \cs_{t+1} = f(\cs_t, u_t), \quad & t\in 0,\dots,T-1,\\
    & \ts_t = {\FK}(\cs_t), \quad & t\in 0,\dots,T, \nonumber
\end{align}
where $f$ is the (unknown) system dynamics, ${\FK}$ is the forward kinematics function, and $\loss$ is a loss function, such as the squared distance $\|\ts_t - \ts_g\|^2$.


\subsection{Reinforcement Learning and Guided Policy Search}\label{ssec:gps}
Reinforcement learning (RL) \cite{sutton1998reinforcement} methods solve the problem in \eqref{eq:prob_def} by interacting with the system, and learning from trial and error. In particular, policy search methods \cite{kober2013reinforcement,deisenroth2013survey} search for the parameters $\theta$ of a reactive parametrized policy $\pi_\theta(u_t|\cs_t)$ such that selecting actions according to $\pi$ minimizes the loss in \eqref{eq:prob_def}.
We focus on the guided policy search (GPS) method \cite{levine2014learning,levine2015learning} -- a model based RL algorithm. However, our approach for combining RL with motion planning can be used with any policy search algorithm. 

GPS is comprised of two main components: trajectory optimization and supervised learning. The trajectory optimization component solves problem \eqref{eq:prob_def} for a fixed set of initial states $\{\cs_0^i\}$ by learning time dependent models $\hat{f}^i(\cs_t, u_t)$ of the system dynamics for each initial state, and computing the optimal policy using an iterative LQG (iLQG) method. The iLQG method optimizes a predefined loss of the form $\sum_t \loss(\ts_t, \ts_g)$.
The second component learns a state-dependent but time-independent neural network (NN) policy $\pi_\theta(u|\cs)$ using supervised learning, by fitting the NN policy to match the actions of the trajectory optimizer from the set of starting positions. A main ingredient in GPS is an additional loss term for the trajectory optimization algorithm that drives the computed solution to be feasible under the NN policy, thereby assuring that the learned policy performs as expected. We refer to \cite{levine2014learning,levine2015learning} for a full description of the method.

\subsection{Motion Planning}\label{ssec:motion_planning}
In robot motion planning \cite{lavalle2006planning,latombe2012robot}, a path between two robot configurations that avoids obstacles is computed. Let $\robot$ and $\obstacles$ denote the geometric description of the robot and obstacles, respectively, and let  $\cspace$ denote the robot configuration space. The free space is defined as $\cfree \doteq \left\{ \cs \in \cspace | \robot(\cs) \cap \obstacles = \emptyset \right\}$. Given an initial configuration $\cs_0$ and goal configuration $\cs_g$, motion planning computes a continuous path $\tau:[0,1] \to \cfree$ such that $\tau(0) = \cs_0$ and $\tau(1) = \cs_g$. In this work we assume that the goal is specified in the task space, and motion planning produces a $\tau$ such that $\tau(0) = \cs_0$ and $\FK(\tau(1)) = \ts_g$.

\section{COMBINING MOTION PLANNING AND POLICY SEARCH}
\label{sec:method}

We begin by motivating our approach with an illustrative example in Figure \ref{fig:example}. In this example, the task is to insert a ring of diameter $r'$ onto a peg of diameter $r$ and height $h$. In the initial state $\ts_0$, the ring is on the floor, next to the peg, and the goal state $\ts_g$ is such that the ring is fully inserted. A policy search algorithm such as GPS, which tries to minimize the cost $\sum_t \loss(\ts_t, \ts_g)$, is likely to arrive at a solution that moves the ring horizontally until it touches peg. Such a solution, which is obviously undesirable, is a local minimum of the cost, and much easier to find with random exploration than the correct solution, especially when $h \gg r$. If we know the geometry of the parts in advance, standard motion planning algorithms can easily generate a collision free trajectory that inserts the ring onto the peg. To execute the plan on the robot, however, we need a controller that generates torque commands that track the trajectory. When the \emph{tolerance} $\delta = r' - r$ is small, such a controller requires extremely high precision, which can be too demanding in practice. To see this, consider the case where, due to sensing noise, the ring is approaching the peg with a horizontal mis-alignment. In this case, successful insertion requires to apply horizontal force to re-align the ring, while a naive tracking controller would apply vertical force to minimize the error with the trajectory, and would get stuck outside the peg. 

In the following, we propose an approach that combines the strength of motion planning in exploiting geometric information about the task, with the effectiveness of policy search in handling complex dynamics and noise in the execution / sensing. Our approach is comprised of three components: a policy search cost function that incorporates the motion plan trajectory, an efficient initialization of the policy search algorithm with a traditional tracking controller, and a NN representation that takes as input a motion plan, and can be trained with RL to track it. Before describing each component in detail, we first outline the complete pipeline of our assembly system.

\begin{figure}
    \centering
    \includegraphics[width=\columnwidth]{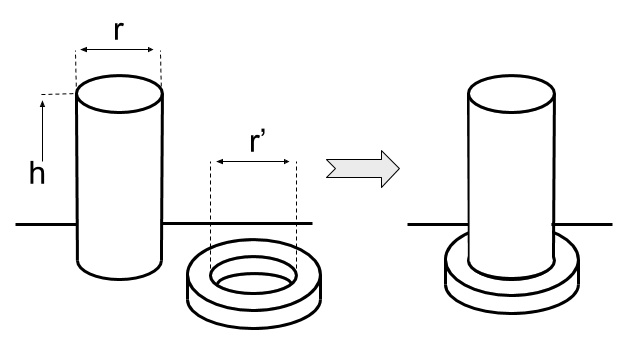}
    \caption{A motivating example. The task is to insert a ring onto a peg. In the starting position (left), the ring is placed on the floor, and the goal is specified by the final position of the ring (right).}
    \label{fig:example}
\end{figure}

\subsection{Robotic Assembly System}\label{ssec:system}
\begin{figure}
    \centering
    \includegraphics[width=\linewidth]{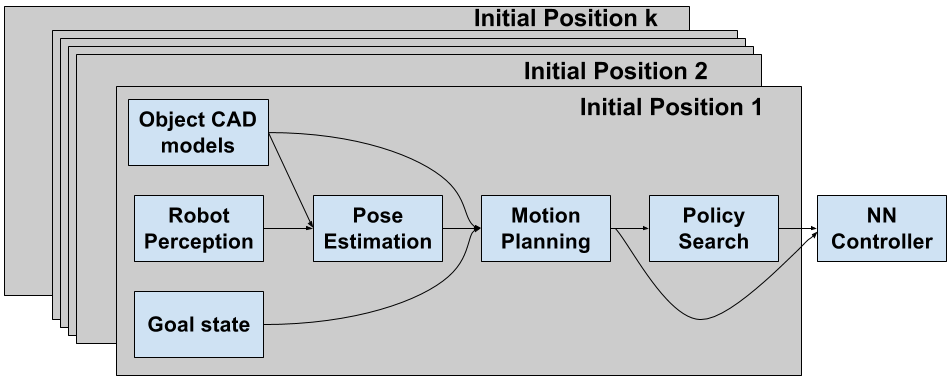}
    \caption{A flow chart of our system. See Section \ref{ssec:system} for details.}
    \label{fig:flow}
\end{figure}

Figure \ref{fig:flow} provides a flow chart of our system.

Given an initial configuration of the parts\footnote{In this work we focus on the assembly of two objects, where one object is grasped and the other is clamped to a fixed position. Our method can be extended to include grasping and assembly of several objects. We defer this to future work.}, we first estimate the pose of the objects, using the geometric CAD data and the robot perception system. For example, in this work we estimated the object pose by using the PR2's binocular vision, and attaching April Tags~\cite{wang2016apriltag} to known positions on the objects, though other methods can be used instead. Given the current pose of the objects, and a known goal position (typically specified in the CAD model), we use an off-the-shelf motion planner to compute a reference trajectory, which is then used to define a cost function, as described in Section \ref{ssec:mp_cost}. This cost function is used by a policy search algorithm, initialized as per Section \ref{ssec:fast_init}, for learning a policy that reaches the goal from the initial position. 

If the same assembly task needs to be performed from varied initial positions, such as random placements of the assembly parts, we can replace the policy search for a new initial position by learning a neural network controller for the task. For training, the process described above is repeated several times with different initial positions but with the same goal position, to obtain a set of policies and motion plans. The motion plans, and the trajectories executed by using the policies are used as data for training a NN controller that follows the reference trajectory using the GPS algorithm~\cite{levine2015learning}, with the NN described in Section \ref{ssec:NN}. After training, the NN controller can replace the policy search for performing the task with a new initial position.

\subsection{A Motion Planning-Based Cost Function}\label{ssec:mp_cost}
Our main observation is that the geometric motion plan can be used to `guide' the policy search onto the correct trajectory to follow, thereby avoiding undesired trajectories that are only locally optimal with respect to the RL cost. We therefore propose a cost function that `tracks' the motion plan, treating it as a reference trajectory. 

Given a manipulation problem as defined in \eqref{eq:prob_def}, we run motion planning to obtain a path $\tau$, where $\tau(0) = \cs_0$ and $\FK(\tau(1)) = \ts_g$ (cf. Sec \ref{ssec:motion_planning}). We then discretize the continuous $\tau$ into $T=\ttraj + \tfinal$ reference points in task space $\rtraj_1,\dots,\rtraj_T$, such that the first $\ttraj$ points are equally spaced on $\tau$, and the last $\tfinal$ points are on the goal,
\begin{equation*}
    \rtraj_t = \begin{cases}
    \FK(\tau(\frac{t}{\ttraj})), & 1\leq t\leq \ttraj,\\
    \ts_g, & \ttraj < t \leq T,
    \end{cases}
\end{equation*}
and propose the following cost function,
\begin{equation}\label{eq:mp_loss}
    \lossmp(\ts_1,\dots,\ts_T) = \sum_{t=1}^{T} \loss(\ts_t, \rtraj_t).
\end{equation}

The cost $\lossmp$ is different from the cost in the original problem \eqref{eq:prob_def}. Thus, in principle, a solution that minimizes $\lossmp$ is not necessarily a solution for \eqref{eq:prob_def}. However, for large enough $T$ and $\tfinal$, the weight of $\loss(\ts_t, \ts_g)$ overwhelms the terms in $\lossmp$, and therefore, trajectories that reach the goal $\ts_g$ would have lower cost than trajectories that do not. Thus, we are guaranteed that the reward shaping \emph{is aligned} with the original problem. 

In this work, we focus on the iLQG algorithm for policy search~\cite{levine2014learning}. 
As we show in our experiments, iLQG with the cost $\lossmp$ performs significantly better than iLQG with a standard cost function $\sum_t \loss(\ts_t, \ts_g)$, by effectively steering the search away from locally optimal solutions.


\subsection{Warm Starting Policy Search with a Motion Plan}\label{ssec:fast_init}
In addition to guiding policy search via the cost function above, we can further exploit the motion plan for warm-starting policy search with a trajectory tracking controller. 

Recall that policy search optimizes a policy $\pi_\theta(u_t|\cs_t)$. Typically, the optimization is initialized with an uninformed policy by randomly selecting $\theta$. However, since in our case we know the motion plan, we can warm-start the search from a policy that acts as a trajectory tracking controller for the motion plan. For example, a proportional-derivative (PD) controller.
Learning the parameters $\theta$ that produce such a controller can be done using supervised learning. 

For the specific case of the iLQG algorithm~\cite{levine2014learning}, however, a more direct approach can be applied.
The output of one iteration of iLQG is a linear control policy. If the state observation includes positions and velocities, a PD controller can also be represented as a linear control policy. Therefore, we can directly replace the first iLQG iteration with a linear controller that applies PD control for tracking the motion plan, thereby warm-starting the algorithm with a well-informed policy.

In our experiments, warm-smarting has significantly reduced the sample complexity of iLQG, allowing us to obtain performant controllers with relatively few task rollouts.



\subsection{A Trajectory-Tracking NN for Generalizing across Task Configurations}\label{ssec:NN}
The cost function and initialization procedure described above consider a specific motion plan, and can therefore be applied for learning the policy for a single task configuration, where by configuration we refer to the initial and final placement of the objects. Indeed, in our work the cost and initialization were used within iLQG, which optimizes the policy from a specific initial state.

In this section we consider the generalization capabilities of our approach, and ask: can we exploit the motion planning component in our method for learning policies that generalize to different task configurations? In particular, we wish to train the policy on a set of configurations, and then be able to solve the task from new configurations that are similar to -- but not exactly alike -- the training configurations.

In principle, the GPS algorithm allows us to do exactly that, by using supervised learning to train a neural network policy that imitates the policies computed by iLQG from a set of training configurations.
However, previous applications of GPS~\cite{levine2014learning,levine2015learning} use a standard multilayer perceptron (MLP; \cite{Goodfellow-et-al-2016}) to represent the policy, which does not make use of the information in the motion plan. Such a representation can generalize well when the task is simple enough that interpolation between the training conditions is sufficient to solve it, or when there is enough training data for essentially `covering' the entire distribution of possible configurations.
However, for tasks that require complex motions and/or have widely varying configurations, we require stronger generalization capabilities.

We next propose a NN architecture that accepts the reference trajectory as additional input and learns to use this information to solve the task in a manner that generalizes naturally to unseen test configurations. We describe the computational graph first and explain the motivation for the design after.

\begin{figure}
   \centering
  \centering
    \noindent
  \begin{subfigure}[b]{0.4\textwidth}
    \includegraphics[width=\textwidth]{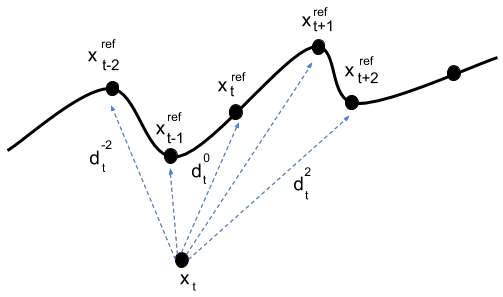}
    \caption{}
  \end{subfigure}
  
  \begin{subfigure}[b]{0.4\textwidth}
    \includegraphics[width=\textwidth]{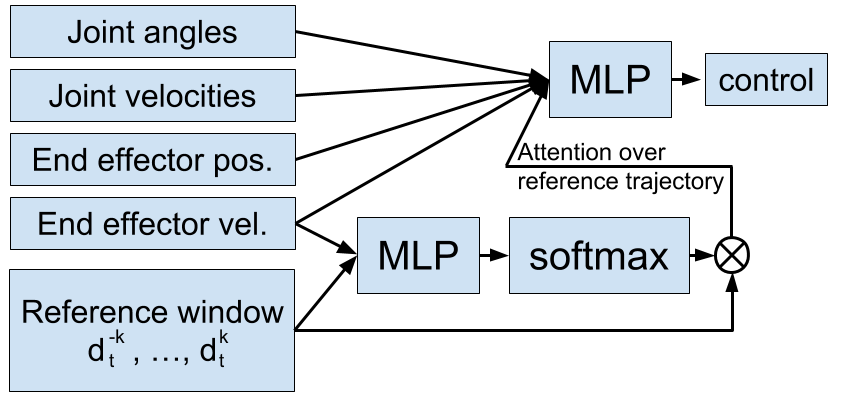}
    \caption{}
  \end{subfigure}
    \caption{A neural network controller with reference trajectory input. (a): The reference trajectory is centered around the current task space pose and cropped to a window around the current timestep. (b): Neural network structure: the reference trajectory window goes through a soft attention, then fed as input along with the state to an MLP for action selection.}
    \label{fig:nn_architecture}
\end{figure}

Our network is depicted in Figure \ref{fig:nn_architecture}. The input to the network consists of the joint angles and angular velocities $\cs_t, \dot{\cs}_t$, end-effector positions and velocities $\ts_t, \dot{\ts}_t$, reference trajectory $\{\rtraj_t\}_{t=1}^T$, and current timestep $t$.
The network first indexes into the reference trajectory to select only the reference points within a time window centered on $t$. We fix a positive integer $k$ (we use $k = 5$ in all our experiments) and select $s_t^i = \rtraj_{\operatorname{clip}(t+i, 1, T)}$ for $i = -k, \dots, k$.
Then we center the points about the current end-effector position by subtracting it out, to obtain a set of directions $d^i_t = s^i_t - \ts_t$.
These directions, as well as the end-effector velocities $\dot{\ts}_t$, are fed into an MLP that outputs a probability distribution $w^{-k}_t, \dots, w^k_t$ over the $2k+1$ directions via the softmax function.
A single attended direction $d^a_t$ is then given by a weighted sum of these:
\[d^a_t = \sum_{i=-k}^{k} w^i_td^i_t.\]
The action $u_t$ is computed by another MLP as a function of the state and this attended direction:
\[u_t = \textsc{mlp}(\phi_t, \dot{\phi}_t, x_t, \dot{x}_t, d^a_t).\]

Using only timesteps $t-k, \dots, t+k$ of the reference trajectory reduces computational load compared to using the whole trajectory, and only local information is needed to determine which direction to move in. In particular the network does not suffer from the issues plaguing reactive policies detailed earlier because global information is encoded into the motion plan.

Subtracting the current end-effector positions from the reference points changes the coordinate system so that the network need only care about the direction to the path it should be following, rather than the absolute position of the path, which is less informative.
During training, this transformation effectively normalizes the data, thereby simplifying the learning process.

Taking a linear combination where the weights are given by a softmax is known as \textit{soft attention}~\cite{xu2015show}, since most of the weights will be roughly zero and thus the result will be approximately equal to one of the directions.
The attentional MLP incorporates velocity information to know how far ahead or behind to look in the given reference data.

Finally, the action is computed as a function of the current state and the direction towards a nearby point on the trajectory. The network learns which direction to select.


\section{EXPERIMENTS}
\label{sec:experiments}
We now report our experimental results. We encourage the reader to view the accompanying video.\footnote{\url{https://youtu.be/pIzuPL5yhcg}}

In our experiments, we set out to investigate the following questions.
\begin{enumerate}
    \item Does our method of combining motion planning with policy search improve upon using each separately?
    \item Can our method effectively solve challenging assembly problems?
\end{enumerate}

\begin{figure}
 \centering
  \begin{subfigure}[b]{0.22\textwidth}
    \includegraphics[width=\linewidth]{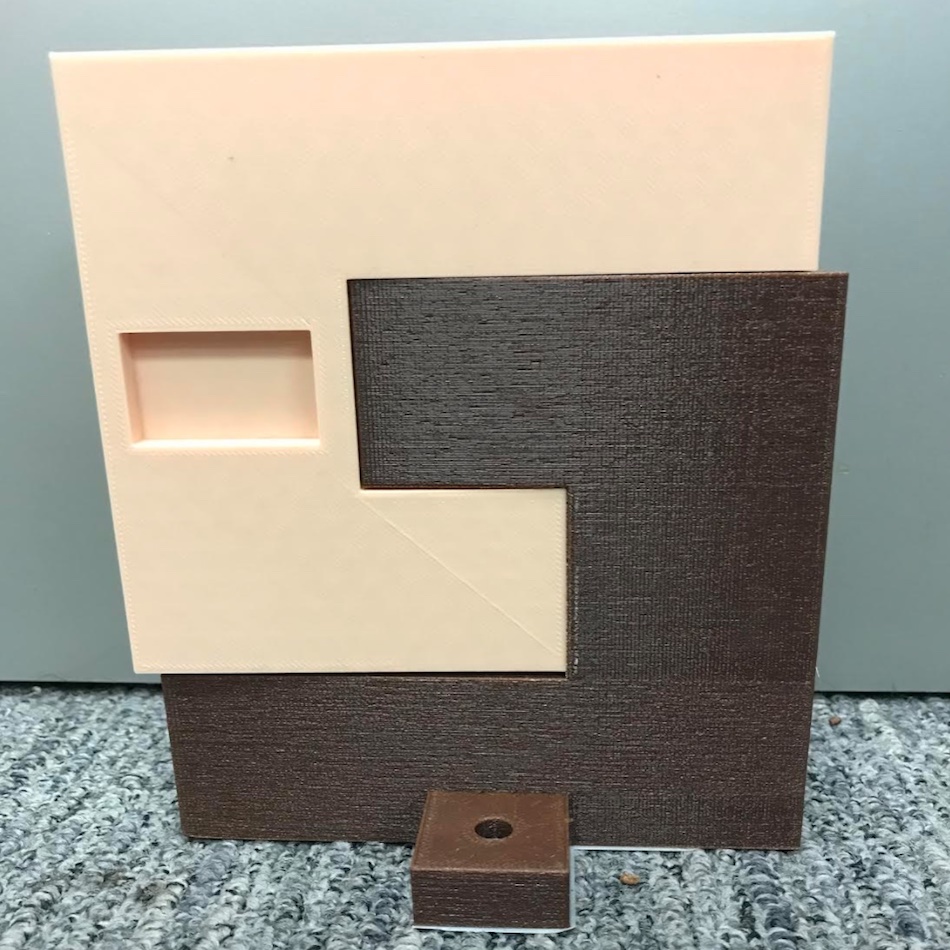}
    \caption{U Shape Assembly Task}
  \end{subfigure}
  \begin{subfigure}[b]{0.22\textwidth}
    \includegraphics[width=\linewidth]{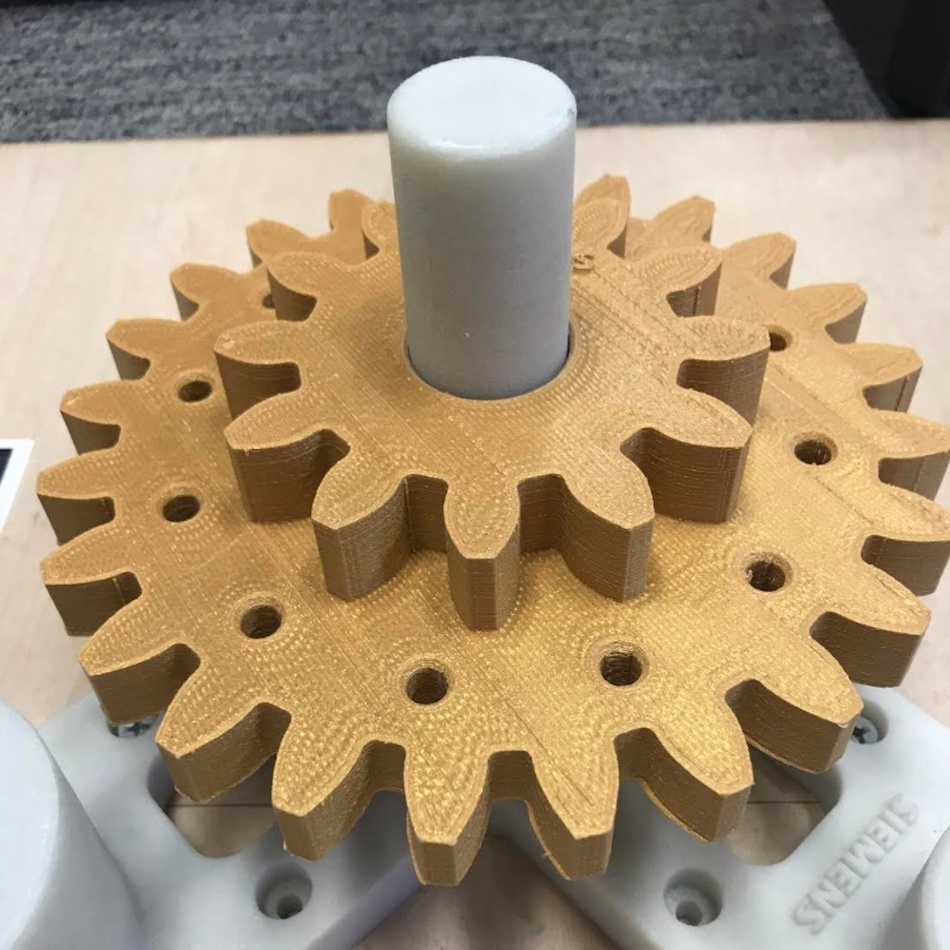}
    \caption{Gear Assembly Task}
  \end{subfigure}
  
  \begin{subfigure}[b]{0.20\textwidth}
    \includegraphics[width=\linewidth]{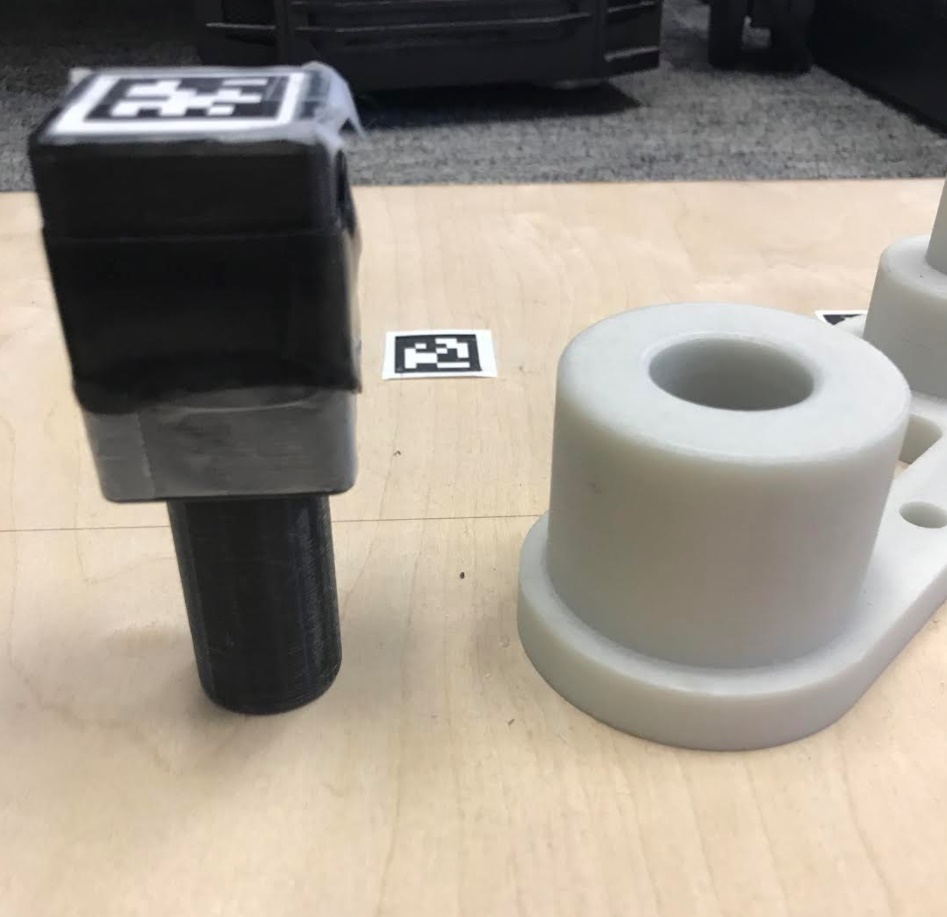}
    \caption{Peg Insertion Task}
  \end{subfigure}
\caption{Assembly tasks in our experiments.}
    \label{fig:ushapegoal}
\end{figure}

We consider the following set of assembly tasks, depicted in Figure \ref{fig:ushapegoal}. These tasks demonstrate the typical difficulties of realistic assembly problems -- tight fitting objects with little tolerance ($1$ mm or less in all tasks), and geometric shapes that require non-trivial motion trajectories for a successful task completion. The parts were 3D printed out of PLA filament using a Type A Series 1 Pro machine, and were not sanded or otherwise post-processed, resulting in an additional high friction challenge. In all tasks, one object is mounted to the table, while the other is initially placed in a random position and orientation on the table\footnote{In this work we do not consider the grasping problem. Thus, each task starts when the object is grasped at a manually defined grasping position.}.
\paragraph{U Shape Assembly Task}
Two U shaped objects need to be interlinked.

\paragraph{Gear Assembly Task}
A gear with a circular aperture of diameter 25mm needs to be fully inserted onto a circular shaft. The shaft height is 67mm, and the gear height is 28mm.

\paragraph{Peg Insertion Task}
A cylindrical peg needs to be fully inserted into a cylindrical hole, of diameter 25mm. The peg length is 46mm.

Tasks (b) and (c) are parts of a full gear assembly task, designed by our collaborators at Siemens as a challenging, realistic problem for robotic assembly.
The CAD files of the objects have been made publicly available\footnote{See \url{http://www.usa.siemens.com/robot-learning}} to facilitate future benchmarking of assembly algorithms.

\textbf{Task Setup Details}

The experiments were done using Willow Garage's robot PR2. For pose estimation, we used AprilTags~\cite{wang2016apriltag}, recorded using the narrow stereo camera on the head of the PR2. To improve pose estimation accuracy, we also recorded the goal position by manually moving the PR2 arm to the goal, and calculating the forward kinematics. The difference between the forward kinematics pose and April tags pose at the goal was later added as an offset to the motion plan. This extra manual step can be avoided by using more accurate pose estimation, which we defer to future work.

For motion planning, we used OMPL~\cite{sucan2012the-open-motion-planning-library}, and the RRTConnect algorithm, which samples both starting and ending states and grows random trees from both. 

For policy search, we used the GPS code implementation~\cite{fzftm-gpsi-16}, and modified the cost function, initialization, and neural network structure according to Section~\ref{sec:method}. To further automate the learning process, 
we also learned reset controllers using iLQG, as described in~\cite{han2015learning}. This was important since learning to pull out the parts from a partly assembled state can require a non-trivial control policy. The GPS implementation sends torque control commands at 20Hz, which are repeated on the robot real time controller at 1KHz. We compare our method to the standard MoveIt!~\cite{moveit} controller, which tracks a joint trajectory at 1KHz. Note that the finer granularity of the MoveIt! controller gives it an advantage over our controller, but in practice the 20Hz controller is sufficient to solve the tasks considered here.

The policy search was warm-started with a joint-tracking PD controller of the form
\[u_t = G[K_p(\phi_t - \phi^*_t) + K_d(\dot\phi_t - \dot\phi^*_t)]\]
where $\phi_t, \dot\phi_t$ are the current joint positions and velocities at time $t$; $\phi^*_t, \dot\phi^*_t$ are the target joint positions and velocities at time $t$ as determined by the motion plan; $G$ is a diagonal gain matrix proportional to the moments of inertia of the PR2's arm links,  from~\cite{fzftm-gpsi-16}.
The values $K_p = 50$ and $K_d = 9$ were chosen manually but kept fixed across all experiments.


\begin{figure}
 \centering
    \noindent
  \begin{subfigure}[b]{0.5\textwidth}
    \includegraphics[width=\linewidth]{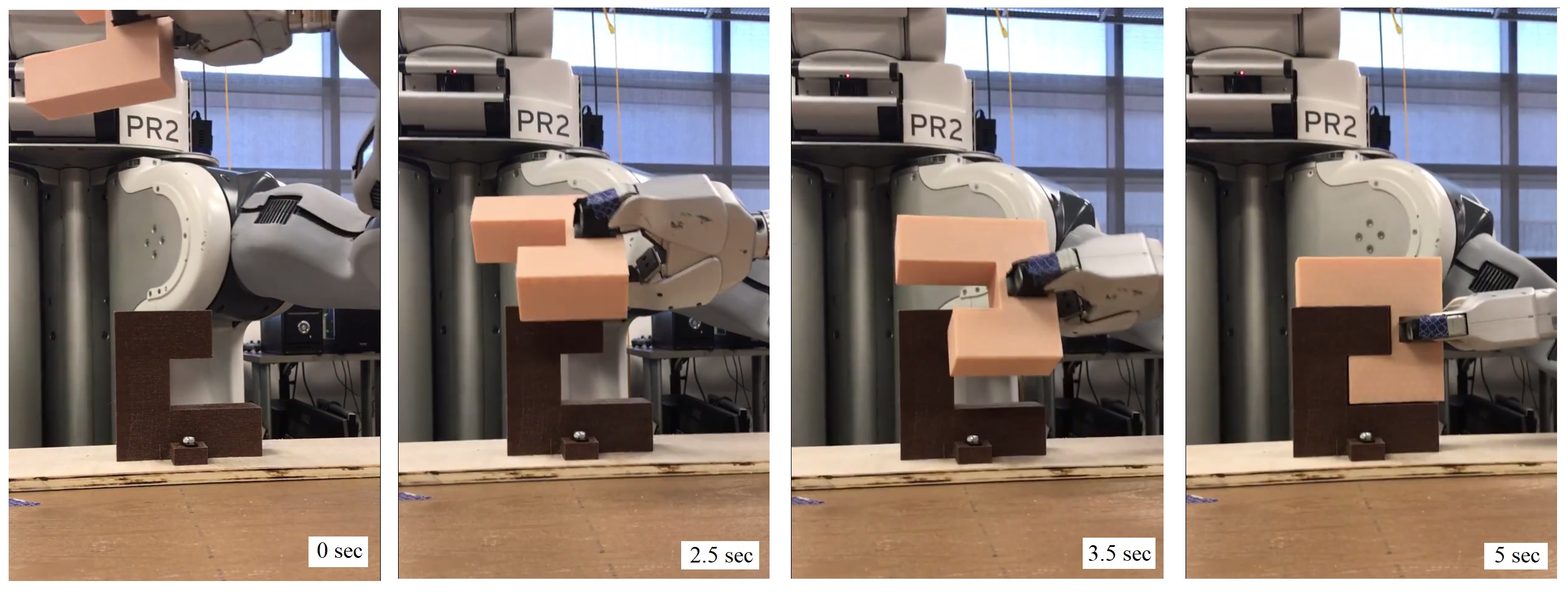}
    \caption{U Shape Assembly Task}
  \end{subfigure}
  
  \begin{subfigure}[b]{0.5\textwidth}
    \includegraphics[width=\linewidth]{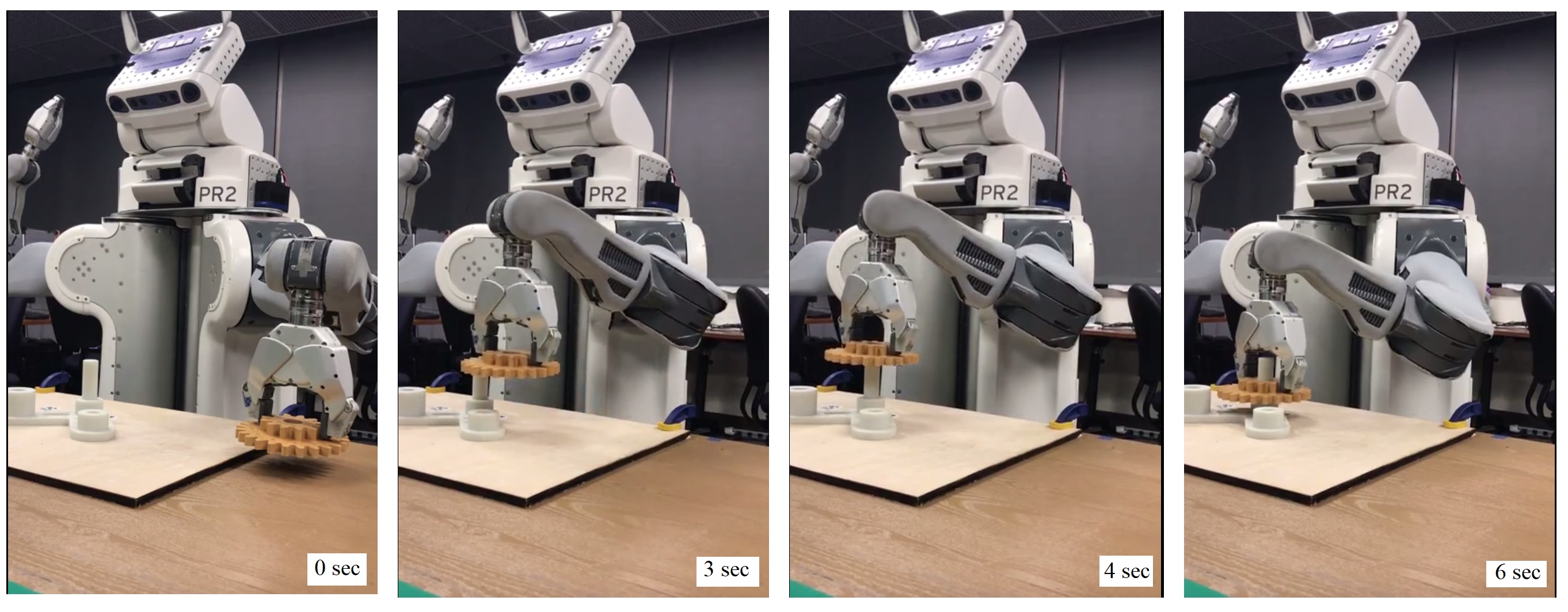}
    \caption{Gear Assembly Task}
  \end{subfigure}
  
  \begin{subfigure}[b]{0.5\textwidth}
    \includegraphics[width=\linewidth]{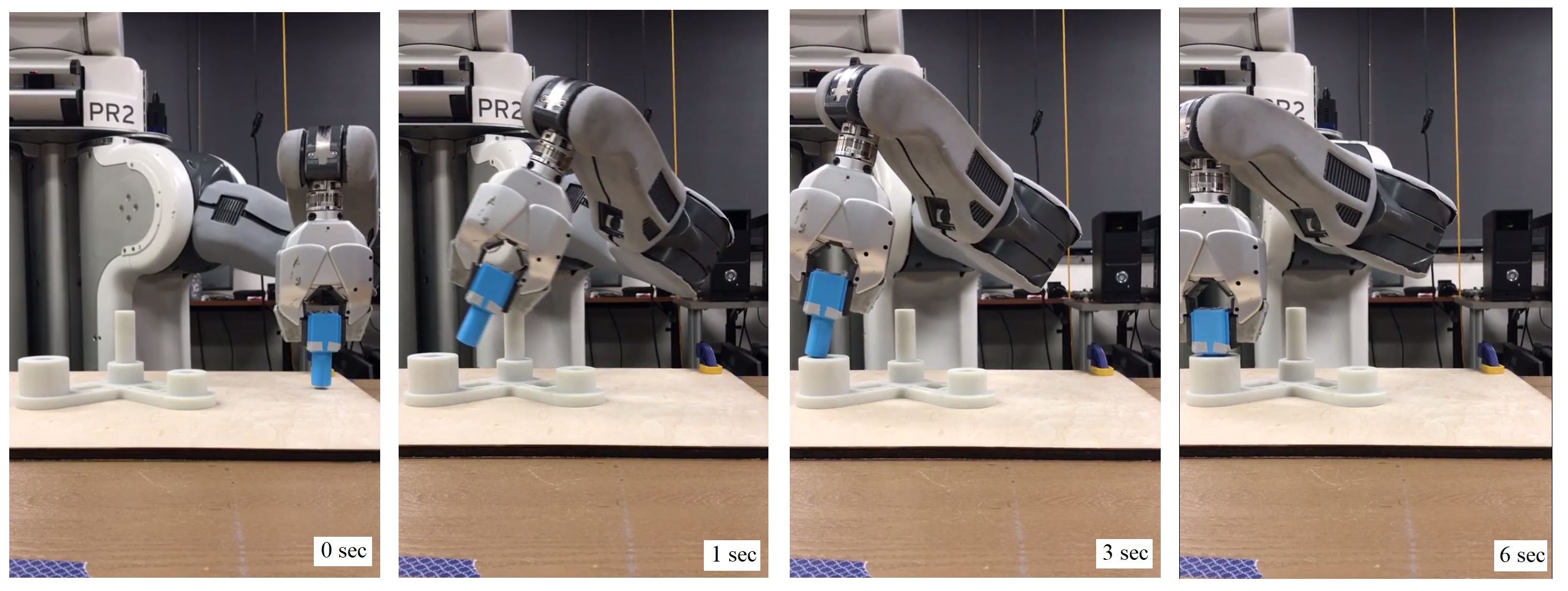}
    \caption{Peg Insertion Task}
  \end{subfigure}
\caption{PR2 successfully assembling the objects in our experiments from the `hard' initial positions.}
    \label{fig:collage}
\end{figure}

\subsection{Evaluating Policy Search with Motion-Planning Cost}
Our first experiment evaluates the cost function introduced in Section \ref{ssec:mp_cost}. For this experiment, We use the iLQG policy search algorithm in~\cite{fzftm-gpsi-16}, and compare our motion planning based cost function with the standard cost function of the form $\sum_t \loss(\ts_t, \ts_g)$. As proposed in \cite{levine2015learning}, we chose $\loss$ to be $\|\ts_t - \ts_g\|^2 + \log(\alpha + \|\ts_t - \ts_g\|^2)$. We also compare to the MoveIt!~\cite{moveit} controller, an off-the-shelf controller for tracking a motion plan, which was given the same motion plan as our algorithm.
For each task, we evaluate the success rate of the policy from two fixed starting positions. The `easy' initial position is such that a straight line motion to the goal can succeed (e.g., the peg placed right above the hole), while the `hard' initial position requires a more complex movement. We define success as reaching the goal position to within less than $2$mm, which corresponds to a full insertion of the objects.
Our results are presented in Table \ref{tab:iLQG}. Clearly, the combination of motion planning with RL resulted in significantly better controllers. The vanilla iLQG method was only able to succeed at the peg insertion task when the peg was placed directly above the hole. The standard MoveIt! controller also failed in all tasks, since the state estimation is not accurate enough to precisely follow the motion plan and avoid contact in the low-tolerance tasks we consider. This is clearly demonstrated in the supporting videos. Combining the motion plan with iLQG, however, resulted in robust policies that can overcome the noise in state estimation and reliably solve the tasks.

In Table \ref{tab:iLQG_samples} we show the number of policy rollouts required for successful task completion. On average, each task required about 30 minutes of robot interaction. 



\begin{table}[]
\centering
\caption{Success rate from fixed positions.}
\label{tab:iLQG}
\begin{tabular}{c|c|c|c|c|c|c|}
\cline{2-7}
                              & \multicolumn{2}{c|}{U Shape} & \multicolumn{2}{c|}{Gear} & \multicolumn{2}{c|}{Peg} \\ \cline{2-7} 
                              & Easy          & Hard         & Easy        & Hard        & Easy        & Hard       \\ \hline
\multicolumn{1}{|c|}{iLQG}    &   0/5            &   0/5           &    0/5         &      0/5       &    5/5         &       0/5     \\ \hline
\multicolumn{1}{|c|}{\textbf{MP+iLQG}} &  \textbf{4/5}             &    \textbf{5/5}          &     \textbf{5/5}        &       \textbf{5/5}      &     \textbf{5/5}        &       \textbf{5/5}     \\ \hline
\multicolumn{1}{|c|}{MP+MoveIt!} &  1/5             &   0/5           &     0/5        &       0/5      &    0/5         &   0/5         \\ \hline
\end{tabular}
\end{table}

\begin{table}[]
\centering
\caption{Sample complexity (number of rollouts) of MP+iLQG from fixed positions.}
\label{tab:iLQG_samples}
\begin{tabular}{c|c|c|c|c|c|c|}
\cline{2-7}
                              & \multicolumn{2}{c|}{U Shape} & \multicolumn{2}{c|}{Gear} & \multicolumn{2}{c|}{Peg} \\ \cline{2-7} 
                              & Easy          & Hard         & Easy        & Hard        & Easy        & Hard       \\ \hline

\multicolumn{1}{|c|}{MP+iLQG} &      75         &      150       &      30       &     30       & 25            &      30      \\ \hline
\end{tabular}
\end{table}

\subsection{Generalization Across Task Configurations}
\begin{figure}
 \centering
    \noindent
  \begin{subfigure}[b]{0.25\textwidth}
    \includegraphics[height=74pt]{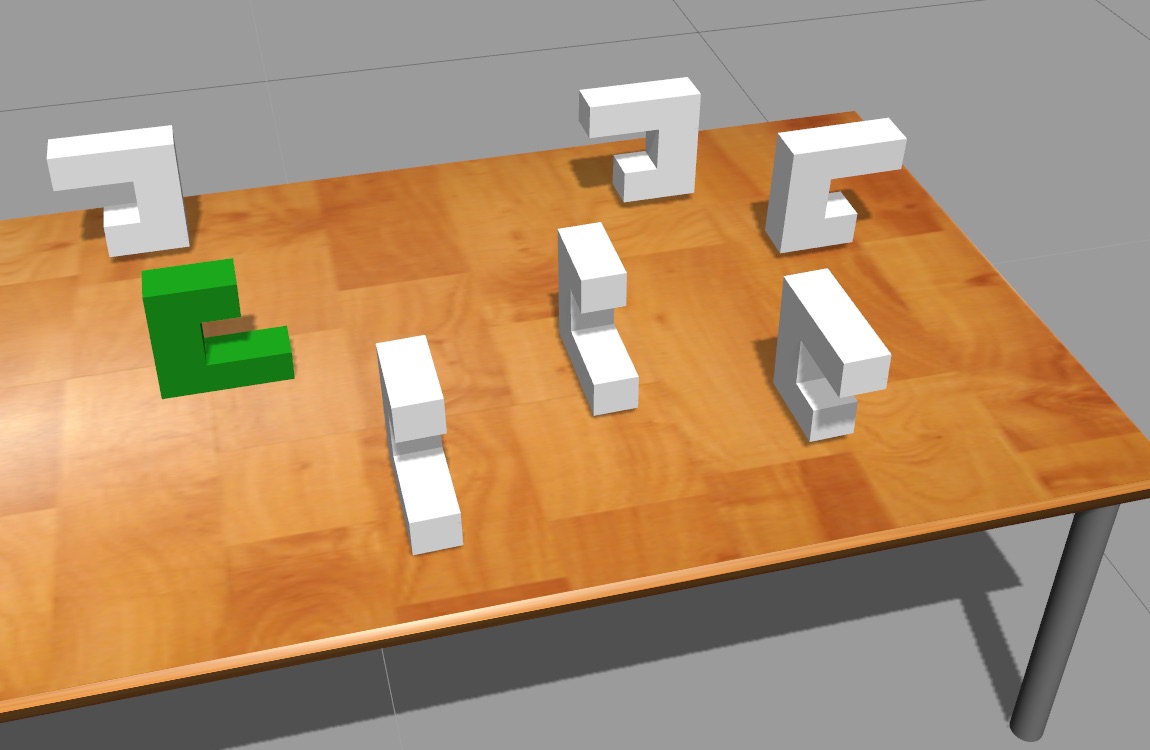}
    \caption{}
  \end{subfigure}
  \begin{subfigure}[b]{0.25\textwidth}
    \includegraphics[height=74pt]{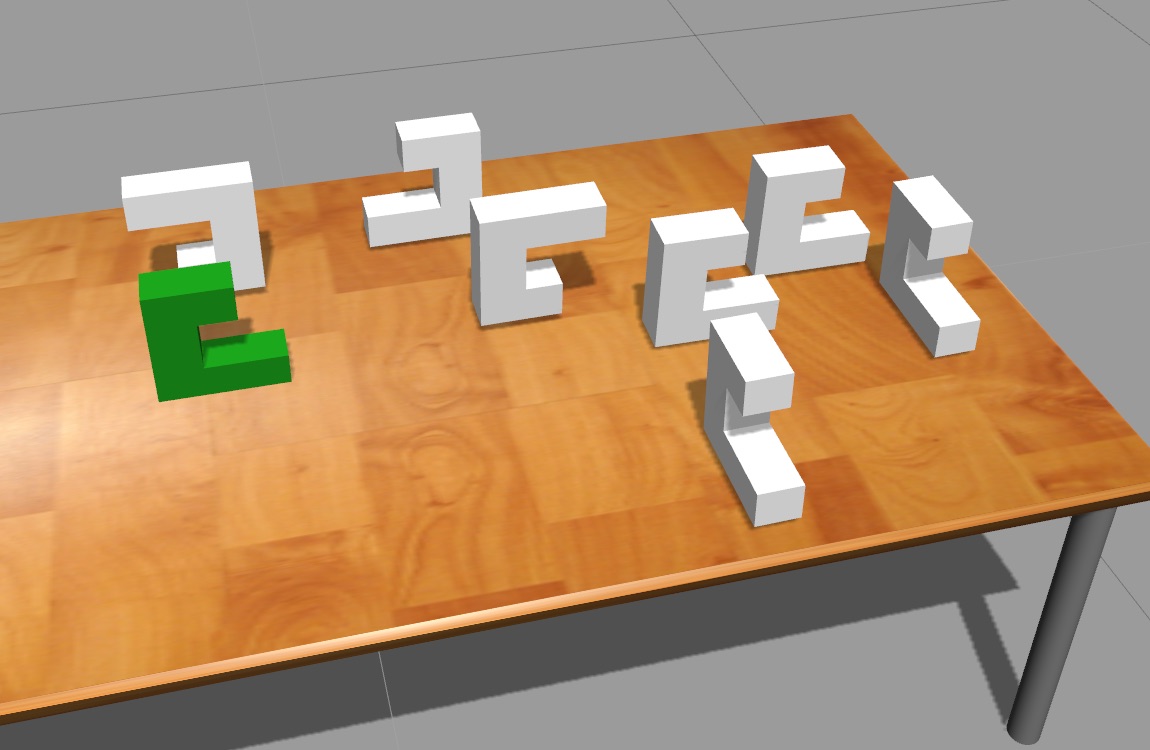}
    \caption{}
  \end{subfigure}
\caption{Initial poses for subsets of the training (top) and test (bottom) sets. The motivation for this setup is an assembly task where the parts are randomly placed in the work area, like a robot working on an assembly line alongside humans.}
    \label{fig:generalization}
\end{figure}

\begin{table}[]
\centering
\caption{Performance on generalization task in simulation}
\label{tab:generalization}
\begin{tabular}{|c|c|c|c|}\hline
& Train & Test & Overall \\\hline
Number of successes & 9 & 18 & 27 \\\hline
Success rate & 0.9 & 0.9 & 0.9 \\\hline
\end{tabular}
\end{table}

To investigate the ability of our proposed network to generalize to unseen test configurations, we prepared a traditional supervised train/test scenario on the U-shape task in simulation.
First a set of 30 initial poses were generated (detailed below) and separated into a training set (10) and a test set (20).
For each of the training poses we trained a controller to solve the task via iLQG with our modified reference cost function.
Then the network was trained to mimic the iLQG controllers using one GPS iteration (equivalent to behavioral cloning).
We used the Adam optimizer \cite{adam} with initial learning rate $10^{-3}$ and $\beta_1 = 0.9$, the other hyperparameters being left as their TensorFlow defaults.

We found it beneficial to pre-train the network (in the same behavioral cloning fashion) on auxiliary data.
This data was also obtained by training iLQG controllers to follow reference trajectories, but these extra trajectories had random initial and goal positions independent of the task being solved. 
This helps because the network needs to learn how to use its attention mechanism over the reference trajectory, rather than just relying on the current state information, as it might do if we only trained it on trajectories that all move towards the same goal position.

At test time, a new reference trajectory is computed for each configuration and passed as input to the policy.
A trial is defined as a success if the end-effector is within 2 mm of the target position. 
Only one rollout is taken for each condition because the system is deterministic.
The results are displayed in Table \ref{tab:generalization}.
Remarkably, the network performs equally well on the training and test sets.

The poses were generated as follows.
We randomly generate $(x, y)$ coordinates uniformly from a rectangular region which roughly corresponds to the area that the robot can reach.
The $z$ position is fixed so that the objects are on (or technically slightly above) the table.
The roll is selected uniformly from $\{0, \pi\}$, the pitch is fixed at $0$, and the yaw is selected uniformly from $\{0, \frac{\pi}{2}, \frac{3\pi}{2}\}$.
Then we attempt to generate a motion plan to this pose to check if it's feasible -- if a plan exists, the pose is added to the set, otherwise we discard and start over.
See Figure \ref{fig:generalization} for a visualization of some of the initial poses generated by this process.

\section{CONCLUSION}
\label{sec:conclusion}
We proposed a method that combines CAD-based motion planning with reinforcement learning, for contact-rich manipulation problems. In our approach, the geometric motion plan guides the reinforcement learning through a special cost function and initialization procedure, resulting in faster and more reliable learning. 
In addition, we proposed a neural network architecture that can imitate the learned solution for tracking the motion plan, thereby generalizing the controller to different task configurations. 

Our method can learn robust controllers in domains that require high precision, even when significant state estimation noise is present. In these challenging domains, state-of-the-art methods were not able to complete the task. Our method is also sample efficient, and requires only minutes of robot time to learn a task. 
We believe that these results provide a promising direction for autonomous assembly that can handle variability in the task/design while being practical for real-world manufacturing domains.


In future work, we will explore the use of additional data that can be retrieved from a CAD file, such as tolerances, process planning, and guidelines, for improving  the reinforcement learning algorithm. Another promising direction is learning policies that work directly with image inputs~\cite{levine2016end}, and can exploit a CAD rendering of a proposed motion plan.


\section{ACKNOWLEDGEMENTS}
\label{sec:acknowledgements}
This work was funded in part by Siemens and by ONR PECASE N000141612723. The authors would like to thank Menglong Guo for 3D-printing the objects used in our experiments.

\appendices

\bibliographystyle{IEEEtran}
\bibliography{refs}

\begin{thebibliography}{10}
\providecommand{\url}[1]{#1}
\csname url@rmstyle\endcsname
\providecommand{\newblock}{\relax}
\providecommand{\bibinfo}[2]{#2}
\providecommand\BIBentrySTDinterwordspacing{\spaceskip=0pt\relax}
\providecommand\BIBentryALTinterwordstretchfactor{4}
\providecommand\BIBentryALTinterwordspacing{\spaceskip=\fontdimen2\font plus
\BIBentryALTinterwordstretchfactor\fontdimen3\font minus
  \fontdimen4\font\relax}
\providecommand\BIBforeignlanguage[2]{{%
\expandafter\ifx\csname l@#1\endcsname\relax
\typeout{** WARNING: IEEEtran.bst: No hyphenation pattern has been}%
\typeout{** loaded for the language `#1'. Using the pattern for}%
\typeout{** the default language instead.}%
\else
\language=\csname l@#1\endcsname
\fi
#2}}

\bibitem{lasi2014industry}
H.~Lasi, P.~Fettke, H.-G. Kemper, T.~Feld, and M.~Hoffmann, ``Industry 4.0,''
  \emph{Business \& Information Systems Engineering}, vol.~6, no.~4, pp.
  239--242, 2014.

\bibitem{durkop2015analyzing}
L.~Durkop, L.~Wisniewski, S.~Heymann, B.~Lucke, and J.~Jasperneite, ``Analyzing
  the engineering effort for the commissioning of industrial automation
  systems,'' in \emph{Emerging Technologies \& Factory Automation (ETFA), 2015
  IEEE 20th Conference on}.\hskip 1em plus 0.5em minus 0.4em\relax IEEE, 2015,
  pp. 1--4.

\bibitem{lavalle2006planning}
S.~M. LaValle, \emph{Planning algorithms}.\hskip 1em plus 0.5em minus
  0.4em\relax Cambridge university press, 2006.

\bibitem{latombe2012robot}
J.-C. Latombe, \emph{Robot motion planning}.\hskip 1em plus 0.5em minus
  0.4em\relax Springer Science \& Business Media, 2012, vol. 124.

\bibitem{donald1993kinodynamic}
B.~Donald, P.~Xavier, J.~Canny, and J.~Reif, ``Kinodynamic motion planning,''
  \emph{Journal of the ACM (JACM)}, vol.~40, no.~5, pp. 1048--1066, 1993.

\bibitem{deisenroth2013survey}
M.~P. Deisenroth, G.~Neumann, J.~Peters, \emph{et~al.}, ``A survey on policy
  search for robotics,'' \emph{Foundations and Trends{\textregistered} in
  Robotics}, vol.~2, no. 1--2, pp. 1--142, 2013.

\bibitem{kober2013reinforcement}
J.~Kober, J.~A. Bagnell, and J.~Peters, ``Reinforcement learning in robotics: A
  survey,'' \emph{International Journal of Robotics Research}, 2013.

\bibitem{levine2015learning}
S.~Levine, N.~Wagener, and P.~Abbeel, ``Learning contact-rich manipulation
  skills with guided policy search,'' in \emph{ICRA}, 2015.

\bibitem{fu2016one}
J.~Fu, S.~Levine, and P.~Abbeel, ``One-shot learning of manipulation skills
  with online dynamics adaptation and neural network priors,'' \emph{IROS},
  2016.

\bibitem{gu2017deep}
S.~Gu, E.~Holly, T.~Lillicrap, and S.~Levine, ``Deep reinforcement learning for
  robotic manipulation with asynchronous off-policy updates,'' in
  \emph{Robotics and Automation (ICRA), 2017 IEEE International Conference
  on}.\hskip 1em plus 0.5em minus 0.4em\relax IEEE, 2017, pp. 3389--3396.

\bibitem{Tamar16}
A.~Tamar, Y.~Wu, G.~Thomas, S.~Levine, and P.~Abbeel, ``Value iteration
  networks,'' in \emph{NIPS}, 2016.

\bibitem{de1991correct}
L.~H. De~Mello and A.~C. Sanderson, ``A correct and complete algorithm for the
  generation of mechanical assembly sequences,'' \emph{IEEE transactions on
  Robotics and Automation}, vol.~7, no.~2, pp. 228--240, 1991.

\bibitem{halperin2000general}
D.~Halperin, J.-C. Latombe, and R.~H. Wilson, ``A general framework for
  assembly planning: The motion space approach,'' \emph{Algorithmica}, vol.~26,
  no. 3-4, pp. 577--601, 2000.

\bibitem{knepper2013ikeabot}
R.~A. Knepper, T.~Layton, J.~Romanishin, and D.~Rus, ``Ikeabot: An autonomous
  multi-robot coordinated furniture assembly system,'' in \emph{Robotics and
  Automation (ICRA), 2013 IEEE International Conference on}.\hskip 1em plus
  0.5em minus 0.4em\relax IEEE, 2013, pp. 855--862.

\bibitem{heger2010assembly}
F.~W. Heger, \emph{Assembly planning in constrained environments: Building
  structures with multiple mobile robots}.\hskip 1em plus 0.5em minus
  0.4em\relax Carnegie Mellon University, 2010.

\bibitem{peters2008reinforcement}
J.~Peters and S.~Schaal, ``Reinforcement learning of motor skills with policy
  gradients,'' \emph{Neural networks}, vol.~21, no.~4, pp. 682--697, 2008.

\bibitem{pastor2009learning}
P.~Pastor, H.~Hoffmann, T.~Asfour, and S.~Schaal, ``Learning and generalization
  of motor skills by learning from demonstration,'' in \emph{Robotics and
  Automation, 2009. ICRA'09. IEEE International Conference on}.\hskip 1em plus
  0.5em minus 0.4em\relax IEEE, 2009, pp. 763--768.

\bibitem{mulling2013learning}
K.~M{\"u}lling, J.~Kober, O.~Kroemer, and J.~Peters, ``Learning to select and
  generalize striking movements in robot table tennis,'' \emph{The
  International Journal of Robotics Research}, vol.~32, no.~3, pp. 263--279,
  2013.

\bibitem{moore2000iterative}
K.~L. Moore and J.-X. Xu, \emph{Iterative learning control}.\hskip 1em plus
  0.5em minus 0.4em\relax Taylor \& Francis, 2000.

\bibitem{longman2000iterative}
R.~W. Longman, ``Iterative learning control and repetitive control for
  engineering practice,'' \emph{International journal of control}, vol.~73,
  no.~10, pp. 930--954, 2000.

\bibitem{bristow2006survey}
D.~A. Bristow, M.~Tharayil, and A.~G. Alleyne, ``A survey of iterative learning
  control,'' \emph{IEEE Control Systems}, 2006.

\bibitem{wang2009survey}
Y.~Wang, F.~Gao, and F.~J. Doyle, ``Survey on iterative learning control,
  repetitive control, and run-to-run control,'' \emph{Journal of Process
  Control}, vol.~19, no.~10, pp. 1589--1600, 2009.

\bibitem{duan2017one}
Y.~Duan, M.~Andrychowicz, B.~Stadie, J.~Ho, J.~Schneider, I.~Sutskever,
  P.~Abbeel, and W.~Zaremba, ``One-shot imitation learning,'' \emph{arXiv
  preprint arXiv:1703.07326}, 2017.

\bibitem{ng1999policy}
A.~Y. Ng, D.~Harada, and S.~Russell, ``Policy invariance under reward
  transformations: Theory and application to reward shaping,'' in \emph{ICML},
  vol.~99, 1999, pp. 278--287.

\bibitem{ng2000algorithms}
A.~Y. Ng, S.~J. Russell, \emph{et~al.}, ``Algorithms for inverse reinforcement
  learning.'' in \emph{Icml}, 2000, pp. 663--670.

\bibitem{abbeel2004apprenticeship}
P.~Abbeel and A.~Y. Ng, ``Apprenticeship learning via inverse reinforcement
  learning,'' in \emph{ICML}.\hskip 1em plus 0.5em minus 0.4em\relax ACM, 2004,
  p.~1.

\bibitem{ziebart2008maximum}
B.~D. Ziebart, A.~L. Maas, J.~A. Bagnell, and A.~K. Dey, ``Maximum entropy
  inverse reinforcement learning.'' in \emph{AAAI}, vol.~8.\hskip 1em plus
  0.5em minus 0.4em\relax Chicago, IL, USA, 2008, pp. 1433--1438.

\bibitem{wirth2016model}
C.~Wirth, J.~Furnkranz, G.~Neumann, \emph{et~al.}, ``Model-free
  preference-based reinforcement learning,'' in \emph{30th AAAI Conference on
  Artificial Intelligence, AAAI 2016}, 2016, pp. 2222--2228.

\bibitem{finn2016guided}
C.~Finn, S.~Levine, and P.~Abbeel, ``Guided cost learning: Deep inverse optimal
  control via policy optimization,'' in \emph{International Conference on
  Machine Learning}, 2016, pp. 49--58.

\bibitem{hsu2002randomized}
D.~Hsu, R.~Kindel, J.-C. Latombe, and S.~Rock, ``Randomized kinodynamic motion
  planning with moving obstacles,'' \emph{The International Journal of Robotics
  Research}, vol.~21, no.~3, pp. 233--255, 2002.

\bibitem{karaman2010optimal}
S.~Karaman and E.~Frazzoli, ``Optimal kinodynamic motion planning using
  incremental sampling-based methods,'' in \emph{Decision and Control (CDC),
  2010 49th IEEE Conference on}.\hskip 1em plus 0.5em minus 0.4em\relax IEEE,
  2010, pp. 7681--7687.

\bibitem{xie2015toward}
C.~Xie, J.~van~den Berg, S.~Patil, and P.~Abbeel, ``Toward asymptotically
  optimal motion planning for kinodynamic systems using a two-point boundary
  value problem solver,'' in \emph{Robotics and Automation (ICRA), 2015 IEEE
  International Conference on}.\hskip 1em plus 0.5em minus 0.4em\relax IEEE,
  2015, pp. 4187--4194.

\bibitem{lange2012control}
F.~Lange, M.~Suppa, and G.~Hirzinger, ``Control with a compliant force-torque
  sensor,'' in \emph{Robotics; Proceedings of ROBOTIK 2012; 7th German
  Conference on}.\hskip 1em plus 0.5em minus 0.4em\relax VDE, 2012, pp. 1--6.

\bibitem{lange2013force}
F.~Lange, W.~Bertleff, and M.~Suppa, ``Force and trajectory control of
  industrial robots in stiff contact,'' in \emph{Robotics and Automation
  (ICRA), 2013 IEEE International Conference on}.\hskip 1em plus 0.5em minus
  0.4em\relax IEEE, 2013, pp. 2927--2934.

\bibitem{sutton1998reinforcement}
R.~S. Sutton and A.~G. Barto, \emph{Reinforcement learning: An
  introduction}.\hskip 1em plus 0.5em minus 0.4em\relax MIT press Cambridge,
  1998, vol.~1, no.~1.

\bibitem{levine2014learning}
S.~Levine and P.~Abbeel, ``Learning neural network policies with guided policy
  search under unknown dynamics,'' in \emph{NIPS}, 2014.

\bibitem{wang2016apriltag}
J.~Wang and E.~Olson, ``Apriltag 2: Efficient and robust fiducial detection,''
  in \emph{Intelligent Robots and Systems (IROS), 2016 IEEE/RSJ International
  Conference on}.\hskip 1em plus 0.5em minus 0.4em\relax IEEE, 2016, pp.
  4193--4198.

\bibitem{Goodfellow-et-al-2016}
I.~Goodfellow, Y.~Bengio, and A.~Courville, \emph{Deep Learning}.\hskip 1em
  plus 0.5em minus 0.4em\relax MIT Press, 2016.

\bibitem{xu2015show}
K.~Xu, J.~Ba, R.~Kiros, K.~Cho, A.~Courville, R.~Salakhudinov, R.~Zemel, and
  Y.~Bengio, ``Show, attend and tell: Neural image caption generation with
  visual attention,'' in \emph{ICML}, 2015.

\bibitem{sucan2012the-open-motion-planning-library}
I.~A. {\c{S}}ucan, M.~Moll, and L.~E. Kavraki, ``The {O}pen {M}otion {P}lanning
  {L}ibrary,'' \emph{{IEEE} Robotics \& Automation Magazine}, vol.~19, no.~4,
  pp. 72--82, December 2012, \url{http://ompl.kavrakilab.org}.

\bibitem{fzftm-gpsi-16}
\BIBentryALTinterwordspacing
C.~Finn, M.~Zhang, J.~Fu, X.~Tan, Z.~McCarthy, E.~Scharff, and S.~Levine,
  ``Guided policy search code implementation,'' 2016. [Online]. Available:
  \url{http://rll.berkeley.edu/gps}
\BIBentrySTDinterwordspacing

\bibitem{han2015learning}
W.~Han, S.~Levine, and P.~Abbeel, ``Learning compound multi-step controllers
  under unknown dynamics,'' in \emph{Intelligent Robots and Systems (IROS),
  2015 IEEE/RSJ International Conference on}.\hskip 1em plus 0.5em minus
  0.4em\relax IEEE, 2015, pp. 6435--6442.

\bibitem{moveit}
\BIBentryALTinterwordspacing
I.~A. {\c{S}}ucan and S.~Chitta, ``{M}ove{I}t!'' [Online]. Available:
  \url{http://moveit.ros.org}
\BIBentrySTDinterwordspacing

\bibitem{adam}
\BIBentryALTinterwordspacing
D.~P. Kingma and J.~Ba, ``Adam: {A} method for stochastic optimization,''
  \emph{CoRR}, vol. abs/1412.6980, 2014. [Online]. Available:
  \url{http://arxiv.org/abs/1412.6980}
\BIBentrySTDinterwordspacing

\bibitem{levine2016end}
S.~Levine, C.~Finn, T.~Darrell, and P.~Abbeel, ``End-to-end training of deep
  visuomotor policies,'' \emph{Journal of Machine Learning Research}, vol.~17,
  no.~39, pp. 1--40, 2016.

\end{thebibliography}
 
\end{document}